\documentclass[letterpaper, 10 pt, conference]{IEEEtran}
\usepackage{cite}
\usepackage{amsmath,amssymb,amsfonts}
\usepackage{algorithmic}
\usepackage{graphicx}
\usepackage{textcomp}
\usepackage{tabularx}
\usepackage{xcolor}
\def\BibTeX{{\rm B\kern-.05em{\sc i\kern-.025em b}\kern-.08em
    T\kern-.1667em\lower.7ex\hbox{E}\kern-.125emX}}
\title{\LARGE \bf
Malleable Agents for Re-Configurable Robotic Manipulators
}

\author{Athindran Ramesh Kumar$^{1}$ and Gurudutt Hosangadi$^{2}$
\thanks{$^{1}$Athindran Ramesh Kumar is with Department of Electrical and Computer Engineering, Princeton University, USA,
	        {\tt\small arkumar@princeton.edu}}%
		\thanks{$^{2}$Gurudutt Hosangadi is with Nokia Bell Labs, Murray Hill, NJ, USA,
				{\tt\small gurudutt.hosangadi@nokia-bell-labs.com}}%
				}

\begin{document}
\maketitle
\thispagestyle{empty}
\pagestyle{empty}
\begin{abstract}
Re-configurable robots have more utility and flexibility for many real-world tasks. Designing a learning agent to operate such robots requires adapting to different configurations. Here, we focus on robotic arms with multiple rigid links connected by joints. We propose a deep reinforcement learning agent with sequence neural networks embedded in the agent to adapt to robotic arms that have a varying number of links. Further, with the additional tool of domain randomization, this agent adapts to different configurations. We perform simulations on a 2D N-link arm to show the ability of our network to transfer and generalize efficiently.
\end{abstract}
\section{Introduction}
Robotic arms play a major role in many industrial and home automation applications. Re-configurable robotic arms \cite{paredis1997rapidly} that can adapt to the task at hand is a futuristic goal with high utility. If the number of links, the shape of the links, and the length of the links can be easily reconfigured, a single intelligent robotic arm can perform a variety of tasks. For example, a cleaning robot can use a slimmer and more flexible link to clean narrow spaces and a more extended link for distant spaces such as a ceiling. Similarly, rescue missions also can benefit from re-configurable arms that depend on the task to be performed.

Recently, reinforcement learning \cite{andrychowicz2020learning, kalashnikov2018qt,gupta2016learning,gupta2017learning, finn2015learning, kumar2016learning, rajeswaran2017learning} has made significant progress in robotic arm manipulation and grasping. While these methods produce attractive results, their real-world feasibility is limited by two major challenges: sample inefficiency and the inability to adapt to another domain which could be the real world. There have been many attempts to address the former challenge by either incorporating expert demonstrations \cite{rajeswaran2017learning, zhu2018reinforcement,gupta2016learning,gao2018reinforcement} or by introducing domain knowledge \cite{ramamurthy2019leveraging,de2018end}. On the other hand, the issue of domain adaptation has been addressed to a lesser extent.

In general, an RL agent is trained in one domain but tested in another domain that has some common structure with the training domain. The ability of the agent to transfer efficiently by learning concepts relevant to the test domain during training time is termed \textit{domain adaptation}. Often, the RL agent is trained in a simulator but deployed in the real world and has to transfer without consuming expensive samples from the real world. This concept is termed as Sim2Real and is an instance of domain adaptation. In the specific scenario of robotic arm manipulation, the test domain can differ from the training domain in terms of the dynamics of the arm, the sensor/camera outputs, or the specific structure of the robotic arm, such as the number of links. The former two sources of domain shift are addressed using techniques such as domain randomization \cite{tobin2017domain,andrychowicz2020learning} or by learning transferable representations \cite{gupta2017learning,higgins2017darla,tzeng2017adversarial,xing2021domain}. More generally, re-configurable arms need agents that can adapt to varying shapes of the links, the number of links, the lengths of the links, and dynamics of the links. Such an agent should retain some of the structure of the underlying robot to capture the semantics of the robot environment. For example, the number of recurring structures in such a neural agent is related to the number of links as the proprioceptive state space and action space grow linearly with the number of links. Even if the shape and size of each link are different, there is common semantics about how each link operates. Using modular differentiable structures for re-configurable robotics has been studied very recently \cite{pathak2019learning, whitman2021learning}.

Inspired by this goal, we propose a novel use of a sequence neural network to generate a single RL agent that can adapt to robotic arms with a varying number of links. We achieve this by treating the state and action vectors as a sequence whose length is the number of links in the arm. We show results on simulated 2D N-link arms to demonstrate the ability of a single agent to perform on 2-link, 3-link, and 4-link arms. Further, we use domain randomization with the GRU agent to generate agents that can generalize to re-configurable arms with a wide range of physical characteristics.

In section \ref{sec:rw}, we discuss related work. In section \ref{sec:pr}, we introduce some preliminary ideas relevant to our work. In section \ref{sec:seq}, we propose a sequence neural network approach to train a single agent that can transfer efficiently for robotic arms with a varying number of links. In section \ref{sec:results}, we demonstrate the agent on a simulated 2D N-link arm. In section \ref{sec:conc}, we conclude our work.
\section{Related Work} \label{sec:rw}
\subsection{Re-configurable robotic manipulators}
Deploying re-configurable manipulators that can adapt to tasks both in terms of the underlying hardware and software was first proposed in \cite{paredis1997rapidly}. Methods to search for the configurations suitable for a particular task are explored using AI methods in \cite{ababsa}. Several methods have been proposed to use classical robust control for precise control of re-configurable robots \cite{li2009decentralized,zhu2010decentralized, zhu2013precision, zhao2018model, gao2016output}. The need for robust control stems from the uncertainties underlying the joint dynamics as these robots can be re-configured. Alternatively, the joint dynamics can be explicitly identified using system identification techniques \cite{li2002identification}. Neural networks have also played a vital role in designing controllers for re-configurable manipulators. Classical neuro-adaptive control methods are explored in \cite{zhao2014local} for adapting to varying joint dynamics with decentralized local control. In \cite{mehraeen2010decentralized}, the Hamilton-Jacobi-Bellman (HJB) formulation from the classical control literature is used with a neural network function approximator to generate controllers. With the advent of deep reinforcement learning (DRL), there is a push towards using DRL for robotics, including re-configurable robots.
\subsection{Deep reinforcement learning for robotic manipulation}
Robotic arm manipulation tasks such as reach, pick-and-place, grasping have become a standard benchmark \cite{mujoco} for testing deep reinforcement learning (DRL) algorithms in recent years. Several DRL methods have emerged to solve challenging real-world robotic manipulation tasks with incredible success. For a general review on the topic, refer \cite{DRLreview}. Initial attempts \cite{rajeswaran2017learning} used DRL with expert demonstrations to learn policies in a tractable fashion. Further, using soft off-policy \cite{lillicrap} versions of reinforcement learning methods such as soft Q-learning \cite{softQ} and soft actor-critic \cite{haarnoja2018soft} have enabled tractable sample efficiency on real-world robots. Vision-based robotic manipulation has also been achieved in \cite{kalashnikov2018qt, andrychowicz2020learning}.

From the context of re-configurable robots, there have been several attempts in recent years to develop modular morphable neural structures that capture the structure of the underlying robots. Initial attempts \cite{schaff2019jointly, chen2018hardware} to generalize agents for different robot configurations conditioned the policy on an encoding of robot structure. In \cite{pathak2019learning}, DRL is utilized to design a morphology for the robot, and the authors use a graph-network based policy that captures the underlying structure of the robot. Similarly, \cite{wang2018nervenet, kurin2020my} also use a graph-network policy to control the robot. In \cite{huang2020one}, the use of message passing graph networks is extended to design a policy that can generalize for a wide variety of morphologies for robots. In \cite{whitman2021learning}, a similar modular graph neural network is used along with model-based reinforcement learning and trajectory optimization. Unlike previous work, we use a recurrent network, a natural reduction of graph networks for robotic arms where links are connected in sequence. We use a completely model-free approach along with domain randomization, unlike most previous work. Another close work to ours is \cite{MBRLsoft} which uses model-based reinforcement learning with recurrent networks for soft robots. Unlike our work, the recurrent network is used for the dynamics model, and there is no attempt to generate general-purpose agents for re-configurable robots. While our work has not been scaled to real-world robotics yet, the simple experiments we perform can provide basic intuition on training modular/recurrent neural networks for re-configurable robots.

\subsection{Domain adaptation and Sim2Real}
There are two aspects to domain adaptation and Sim2Real transfer: one is the variability of dynamics between simulation and reality, and the other is the discrepancy in the distribution of sensor inputs between simulation and reality. The former problem has been addressed using domain randomization approaches that train the agent in many environments. In these environments, the physical parameters are randomized, and dynamics are tweaked so that the agent learns to adapt to these changes robustly. Domain randomization has been used for object localization in images in \cite{tobin2017domain} and for manipulating a Rubik's cube with a robotic arm in \cite{andrychowicz2020learning}.
\par
Another aspect of domain adaptation is when the image or sensor inputs to the agent come from a different distribution from the one in which it is trained. In \cite{gupta2017learning}, domain adaptation is performed between environments that have similar semantics, but the structure of the state spaces and action spaces are different. The agent learns to do a simple proxy task in both environments. This task is used to learn an invariant representation that is amenable to both environments. This invariant representation is used to transfer quickly to a different environment. Similarly, in \cite{higgins2017darla,tzeng2017adversarial,xing2021domain}, the approaches learn a representation of visual inputs that are designed to adapt to different domains.  
\par
Some other approaches to transfer between different domains use progressive networks \cite{rusu2017sim} that build upon networks trained in simulation to transfer quickly to reality. Further, in \cite{rajendran2015attend}, an attention mechanism is used to transfer learning from an ensemble of tasks to a new task while using lesser samples. 
\par
In this work, we consider a third aspect of domain adaptation that is specific to a reconfigurable multi-link robotic arm. We introduce a agent that can adapt to robotic arms with various links, flexible lengths, and robust to the impact of varying dynamics. In our approach, the semantics of the N-link arm are embedded into the neural network through a sequence model.
\section{Preliminaries}\label{sec:pr}
\subsection{Markov decision process}
A major component of reinforcement learning is a Markov Decision Process (MDP) which has the components: $s\in \mathcal{S}$ - state space, $a \in \mathcal{A}$ - action space, $r(s,a)$ - reward, a transition function $p(s^{\prime}|s,a)$ and a time horizon $T$ of an episode. The discount factor $\gamma$ is also used optionally even though the finite horizon formulation does not need it. The goal of reinforcement learning is to design a policy $\pi(a|s)$ which maximizes cumulative reward over an episode. 
\subsection{Soft actor-critic}
The soft actor critic \cite{haarnoja2018soft} is an off-policy algorithm that is suitable for continuous action spaces. In this algorithm, there are two neural networks: the actor network and critic network. The actor network tries to learn a policy that maximizes the reward-to-go denoted by $Q(s,a)$ while the critic network learns $Q(s,a)$ by using the experiences in the environment. The soft actor-critic is an off-policy algorithm that stores experiences in a replay buffer and reuses samples to optimize the actor and critic. In order to aid exploration, a soft policy is used which samples from a Gaussian distribution with mean and standard deviation provided by the actor network. The actor network learns a policy $\pi$ by optimizing the following function:
\begin{align*}
    \pi^{\ast}&=\arg \max \mathbb{E}_{\tau \sim \pi} \left[ \sum_{t=0}^{T} \gamma^{t} \left(r(s_{t},a_{t})+\alpha H(\cdot | s_{t})\right)\right]
\end{align*}
The critic network learns the $Q(\cdot,\cdot)$:
\begin{align*}
    Q^{\pi}(s,a)&=\mathbb{E}_{\tau \sim \pi} \left[\sum_{t=0}^{T} \gamma^{t}r(s_{t},a_{t})+\sum_{t=1}^{T} \gamma^{t}\alpha H(\cdot | s_{t}) | s_{0},a_{0}\right]
\end{align*}
where $\alpha$ is an appropriate weighting factor for the entropy term, $\gamma$ is the discount factor and $H$ denotes the entropy of the policy. The critic module maintains two $Q$ networks to avoid maximization bias and updates a separate target $Q$ network by polyak averaging. The $\alpha$ parameter is also optimized to achieve a target entropy.
\subsection{Environment Setup}
\begin{figure}
  \centering
  \fbox{\includegraphics[scale=0.4]{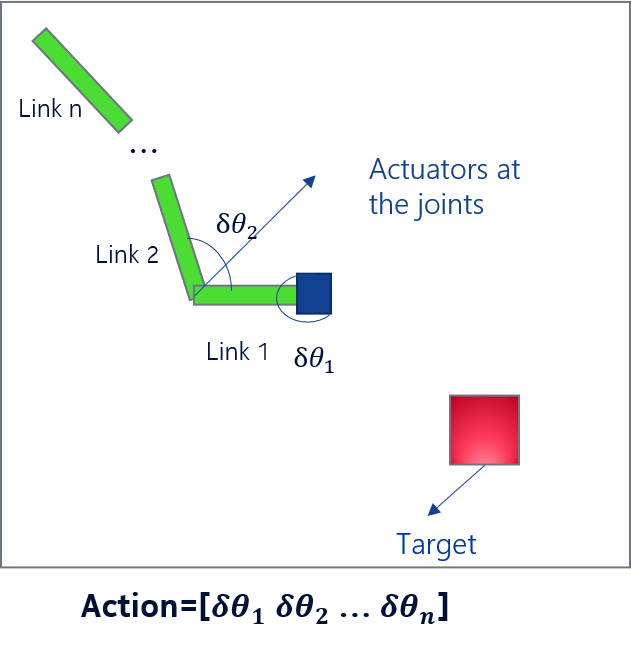}}
  \caption{Simulated 2D N-link arm used for testing}
  \label{fig:4link}
\end{figure}

For our experiments, we envision a re-configurable arm with $n$ links with the length of each link being $l_i$ where $i \in {1,\ldots, n}$. Both $n$ and $l_i$ can be configured to modify the coverage area for the task of reaching a goal. Specifically, we use a 2D robotic arm environment with
$n$ links with $n=2,3,4$ and a continuous state-action space is used for testing our methods. The state in each environment comprises of the positions of the end-points of the links and the distance of each end-point from the goal:
\begin{align}
    s_{t}&=\left[p^{1}_{t},g-p^{1}_{t},p^{2}_{t},g-p^{2}_{t},\dots,p^{n}_{t},g-p^{n}_{t}\right]
\end{align}
where $g$ is the 2D position of the goal and $p^{i}_{t}$ is the 2D position of the end point of link $i$ in the arm at time $t$. All positions are relative to the fixed joint of the arm. This specific structure for the state that includes both $(p^{i}_{t},g-p^{i}_{t})$ is seen to greatly accelerate learning. The goal region is a square target randomly generated in this 2D space. The task for the robotic arm is to reach the goal region. The size of the 2D space is proportionally scaled as the number of links increases. The reward is designed to proportionally decrease with the distance to the goal. Once, the finger of the arm is in the goal for $50$ time steps, a reward of $250$ is provided and the episode ends. The number of time steps for which the finger is present at the goal is appended to the state. This information is needed by the agent to provide information as to whether the task is nearing completion. The actions provided by the reinforcement learning agent are given by:
\begin{align}
	a_{t} &= \left[ \theta_1, \ldots, \theta_n \right]
\end{align} 
where $\theta_i$ is the angular perturbation applied to the servo motor that controls link $i$. A visual description of the N-link arm environment created is shown in figure \ref{fig:4link}.
\section{Architecture of the Agent}\label{sec:seq}
Irrespective of the number of links in the robotic arm, the underlying semantics of the n-link environments are quite similar. We want to design a single agent that can work independent of the number of links in the robotic arm. Further, when used on a new arm with a different number of links, we desire the transfer to be sample-efficient.

\begin{figure}
  \centering
  \includegraphics[scale=0.38]{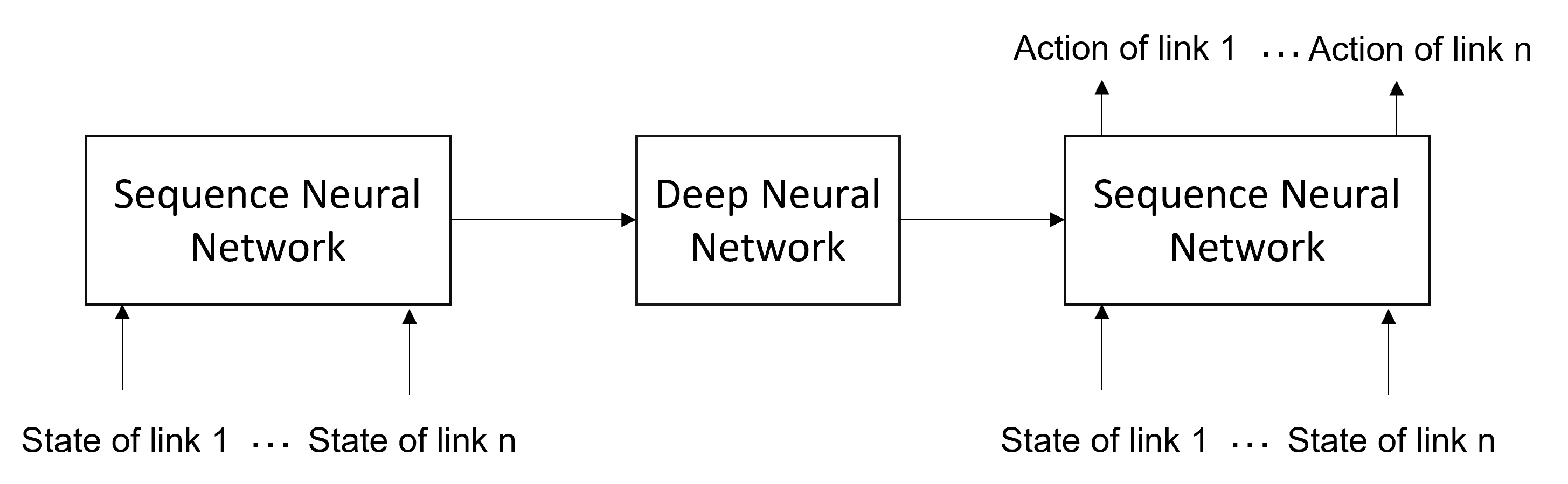}
  \caption{Architecture of the neural network designed for adapting to arms with varying number of links}
  \label{fig:seqnet}
\end{figure}

While our method can be used with any RL algorithm, we choose to use the Soft Actor-Critic (SAC) \cite{haarnoja2018soft}. We design the actor and critic as comprising of sequence neural networks. Specifically, we choose to use a Gated Recurrent Unit (GRU)\cite{cho2014learning} for the sequence network. The actor and critic GRU can process a sequence of states where each individual state is the state of the endpoint of each link in the entire robotic arm. The state of link $i$ is:
\begin{align}
s^{i}_{t}&=\left[p^{i}_{t}, g-p^{i}_{t}\right]    
\end{align}
In figure \ref{fig:seqnet}, we depict the structure of the actor network that comprises of a sequence network to process the states, followed by a deep neural network and a sequence network to provide the actions for each link of the arm. Unlike previous work \cite{huang2020one,pathak2019learning}, there is no decentralized two-way message passing. Instead, there is one sequence network that learns a representation of the states of $n$ links, which acts a context for generating the actions for the $n$ links. The critic network only needs the sequence network to process the states. 

The GRU can transfer some learning to sequences with a different length more easily, thus enabling more sample-efficient transfer between robotic arms with different number of links. We term this recurrent neural network based approach as Rec-SAC henceforth.
\section{Simulation Results}\label{sec:results}
\subsection{Efficient transfer with varying number of links}
\begin{figure}
    \centering
    \includegraphics[scale=0.4]{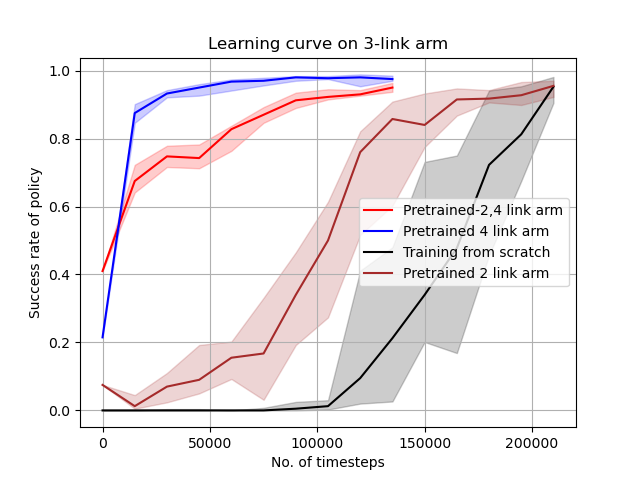}
    \caption{Transfer between arms with different number of links. A Rec-SAC agent trained on a 2-link and 4-link arm can transfer faster on a 3-link arm}
    \label{fig:transfer}
\end{figure}

We train the Rec-SAC method on both a 2-link and 4-link arm simultaneously and test it on a 3-link arm. The GRU of the actor and critic networks are comprised of internal hidden states of size $28$. The deep neural network of the actor and critic are comprised of $4$ layers with $1024$ hidden units. After the training, we tested the agent with $200$ seeds with the initial joint positions and the goal positions randomized, and recorded the success rate. The trained agent achieves $89\%$ success and $92.5\%$ success in the 2-link arm and 4-link arm. This agent can achieve $41\%$ success without further training in the 3-link arm clearly showing that the network has learnt some transferable components during training. Also, the agent trained purely on the 4-link arm and not on both the 2-link arm and 4-link arm simultaneously can provide a zero-shot success rate of $21.5\%$ on the 3-link arm. Both these pre-trained networks can transfer more efficiently than training the agent from scratch on the 3-link arm as depicted in Figure \ref{fig:transfer}. We used the same network for both the training but tuned the hyper-parameters for reasonable performance. Transferring from the pre-trained network is seen to be consistently more efficient. The solid lines correspond to the $50\%$th percentile while the shaded region around the solid line cover $25-75$ percentiles. The training curves are averaged over 10 different seeds.
\par
\begin{figure}
    \centering
    \includegraphics[scale=0.4]{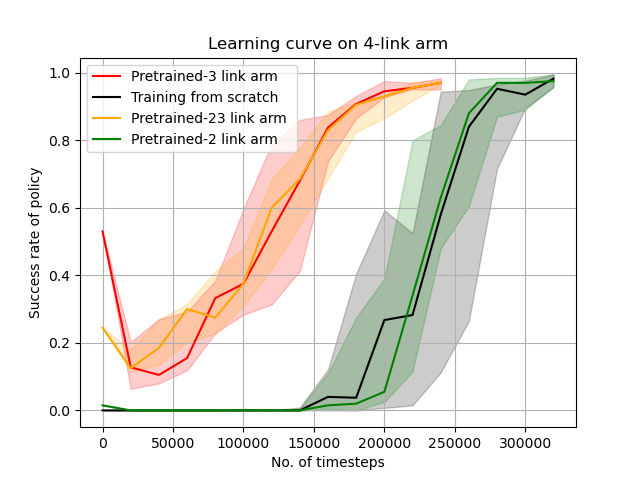}
    \caption{Transfer between arms with different number of links. A Rec-SAC agent trained on a 3-link arm can transfer faster on a 4-link arm}
    \label{fig:transfer2}
\end{figure}

Similarly, the Rec-SAC method is trained only on the 3-link arm environment to get $95\%$ success in the 3-link arm. Without any further training, it can provide $51\%$ success in the 4-link arm. Finally, we see that the training with the pre-trained model is more sample efficient as compared to training from scratch in Figure \ref{fig:transfer2}. In this case, the training from scratch is done with a network with one hidden layer lesser. Otherwise, the training takes even more samples without prior training in other environments.
\par
\begin{figure}
    \centering
    \includegraphics[scale=0.7]{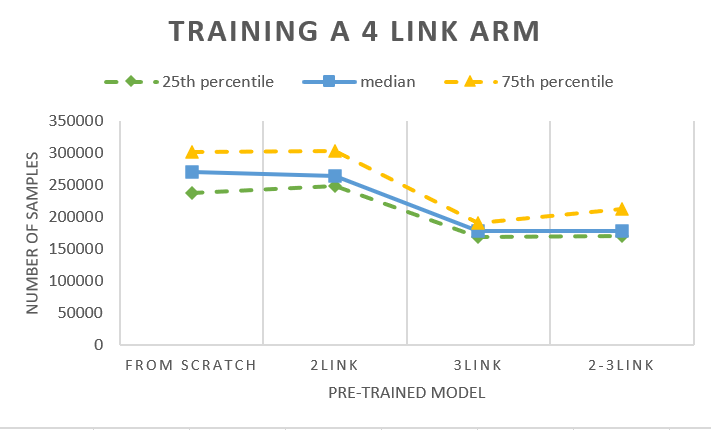}
    \caption{Sample complexity to train a $4$-link arm from scratch compared to using a pre-trained model from other arms to achieve a $90\%$ success rate.}
    \label{fig:4-link-train}
\end{figure}
\begin{figure}
    \centering
    \includegraphics[scale=0.7]{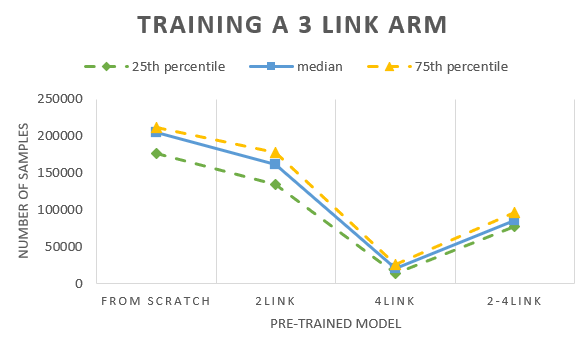}
    \caption{Sample complexity to train a $3$-link arm from scratch compared to using a pre-trained model from other arms to achieve a $90\%$ success rate.}
    \label{fig:3-link-train}
\end{figure}

Additional insight into the adaptability of our Rec-SAC agent can be gained by looking at the sample complexity for training an $N$-link arm using a model pretrained on $N\pm x$ compared to training from scratch to acheive a success rate of $90\%$. Figure \ref{fig:4-link-train} shows that training efficiency for a $N=4$-link is improved by about $35\%$ when using pretrained model on $3$-link. There only a slight improvement of around $2\%$ if a pretrained model on $2$-link is used. Figure \ref{fig:3-link-train} shows similar results when a $3$-link arm is trained from scratch compared to using a pretrained model on arm enviroments with more (specifically $4$) or fewer ($2$) links. When a model pretrained on 4-link arm is used for training the 3-link arm agent, the sample complexity is reduced by nearly a factor of $10$. 

\subsection{Domain Randomization for a Generalizable Agent}
An additional desiderata for our work is the need to develop agents that can transfer to arms with different configurations. Specifically, we are interested in generalizing to varying number of links, different lengths for the links and random sources of noise. To achieve this, we just use our Rec-SAC agent with domain randomization. 

Specifically, we create multiple environments which are variations of the original 4-link and 3-link environments. When an RL agent provides an action ($[\theta_i]_{i \in {1,\ldots,N}}$), the angular perturbation that is actually applied to motor controlling the link may vary due to frictional forces at the joints or small vibrations of the link. We model this noise as comprising of $2$ components: the first component is represented by $\mathcal{U}$ which is uniform in range $[\mathcal{E}_{\theta_l},\mathcal{E}_{\theta_h}]$ while the second component $\mathcal{F}$ represents a non-linear quadratic term given by $\mathcal{F}(\theta_i)=\theta_i - \beta \theta_i^2$\cite{maksim2019learning}.  
The descriptions of the environments are outlined in Table \ref{tab:envtable}.

\begin{table}
\centering
	\begin{tabular}{|c|c|p{3.9cm}|c|c|}
 \hline
	Env & $N$ & $L_i$ & $\mathcal{U}$ & $\mathcal{F}$ \\
 \hline
	1 & 4 & $L_i=L, i \in {1,\ldots,N}$ & No & No \\
\hline
	2 & 4 & $L_1=L_4=L$, $L_2=1.5L$, $L_3=0.5L$ & No & No \\
\hline
	3 & 4 & $L_1=L_4=L$, $L_2=1.5L$, $L_3=0.5L$ & Yes & Yes \\
\hline
	4 & 3 & $L_i=L$, $i \in {1,\ldots,N}$ & No & No \\
\hline
	5 & 3 & $L_1=L$, $L_2=1.5L$, $L_3=0.5L$ & No & No \\
\hline
	6 & 4 & $L_1=L_4=0.3L$, $L_2=L_3=0.7L$ & No & No \\
\hline
	7 & 3 & $L_1=0.5L$, $L_2=1.2L$, $L_3=1.3L$ & No & No \\
\hline
	8 & 4 & $L_1=0.7L$, $L_2=1.3L$, $L_3=1.55L$, $L_4=0.45L$ & Yes & No \\
\hline
    9 & 3 & $L_1=0.5L$, $L_2=1.5L$, $L_3=L$ & Yes & Yes\\
\hline    
\end{tabular}
	\caption{\textbf{Environments} considered to evaluate domain randomization.}
	\label{tab:envtable}
\end{table}

\begin{figure}
  \centering
	\fbox{\includegraphics[scale=0.45]{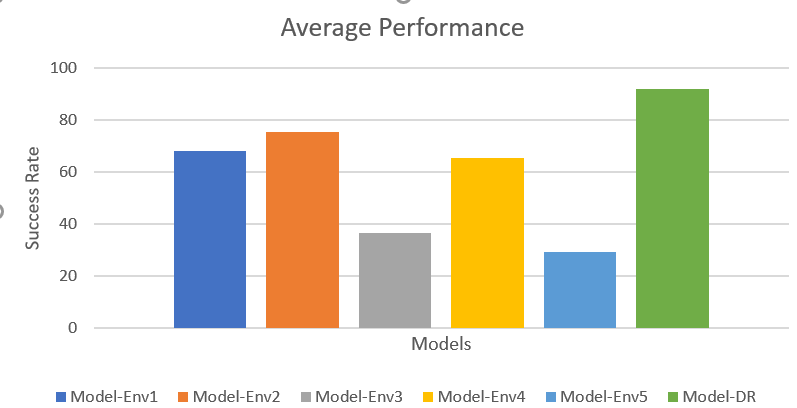}}
   \caption{Average Performance of the different models across environments simulated}
   \label{fig:rate-plot}
\end{figure}

We train individual Rec-SAC agents for each environment and test their ability to generalize to the others. Further, we train a single Rec-SAC agent using randomization on the first five environments during training. We test the performance of this agent on all eight environments. The results of this testing is reported in Table \ref{tab:drtable}. The model trained on a 3-link environment does provide decent success on 4-link arm but cannot work without further training. The vice-versa is also true. Further, Env3 is a harder 4-link environment and has non-linear terms which make the operation somewhat orthogonal to the other 4-link environments. The agent trained with domain randomization on Env 1-5 can generalize quite well and has some reasonable success on all eight environments. The closest competitor is Rec-SAC-Env2 which performs well on all the 4-link environments but is not sufficient on the 3-link environments. The average performance of each model across all environments is shown in Figure \ref{fig:rate-plot}. Thus, our approach of using Rec-SAC with domain randomization can provide generalizable agents that can adapt to a variety of environments.
\begin{table*}
\centering
\begin{tabularx}{0.95\textwidth} { 
  | >{\raggedright\arraybackslash}X 
  | >{\centering\arraybackslash}X
  | >{\centering\arraybackslash}X
  | >{\centering\arraybackslash}X
  | >{\centering\arraybackslash}X
  | >{\centering\arraybackslash}X
  | >{\centering\arraybackslash}X 
  | >{\centering\arraybackslash}X 
  | >{\centering\arraybackslash}X 
  | >{\centering\arraybackslash}X | }
 \hline
	Model& Env 1 (4-link) & Env 2 (4-link)& Env 3 (4-link)& Env 4 (3-link)& Env 5 (3-link)& Env 6 (4-link)& Env 7 (3-link) & Env 8 (4-link) & Env 9 (3-link) \\
 \hline
	Rec-SAC-Env1 & 99.0\%& 93.5\%& 74.5\%&  40.0\%& 45.5\%& 97\%& 36.5\%& 97\% & 29\% \\
  \hline
	Rec-SAC-Env2 & 95.0\%& 97.0\%& 95.5\%& 56\%& 58\%& 99.5\%& 43\%& 96.5\%  & 37.5\% \\
 \hline 
	Rec-SAC-Env3 & 19.0\%& 63.5\%& 91.5\%& 5.5\%& 5\%& 65.0\%& 4\%& 63.5\%  & 13\% \\
 \hline
	Rec-SAC-Env4 & 60.5\%& 57.0\%& 33.5\%& 98.5\%& 90.0\%& 35.0\%& 97.5\%& 33\%  & 83.5\% \\
 \hline
	Rec-SAC-Env5 & 16.5\%& 21\%& 14\%& 56.5\%& 97\%& 12\%& 33.5\%& 7\% & 39.5\% \\
 \hline
	\textbf{Rec-SAC-DR} & 94.5\% & 89\% & 95.5\% & 96.5\%& 95.5\% & 92.5\%& 85\%& 89\% & 86\% \\
\hline
\end{tabularx}
\caption{\textbf{Success rate} of goal capture for the different robotic arm environments. There is one model trained individually on each environment from 1-5 and test on all the others. Further, the GRU agent (Model-DR-GRU) is trained with domain randomization on environments (1-5) and tested on all environments.}\label{tab:drtable}
\end{table*}

\section{Conclusions and Future Work}\label{sec:conc}
A major challenge for deep reinforcement learning is to adapt to domains different from the one it is trained on. This is an important problem which limits application in real-world robotics. Adapting to robotic arms with different dynamics due to wear and tear, variability in contact friction or sensors with different characteristics have been explored in the literature. In this work, we discussed an approach to design an agent that can adapt to robotic arms with varying number of links. Further, this approach provides a mechanism to design an agent that can adapt to re-configurable robots with varying number and type of links.
\par
As future work, it would be interesting to transfer this agent to real robotic arms by performing Sim2Real on arms with different number of links and dynamics. Further, with the flexibility of the malleable agent, it would be possible to use expert demonstrations on a different type of manipulator to accelerate learning on a provided manipulator.
\bibliographystyle{unsrt}
\bibliography{refs}
\end{document}